\documentclass[11pt]{article}

\usepackage[preprint]{acl}

\usepackage{times}
\usepackage{latexsym}

\usepackage[T1]{fontenc}

\usepackage[utf8]{inputenc}

\usepackage{microtype}

\usepackage{inconsolata}

\usepackage{graphicx}

\usepackage{amsmath}
\usepackage{amssymb}
\usepackage{booktabs}
\usepackage{multirow}
\usepackage{enumitem}
\usepackage{makecell}
\usepackage[table]{xcolor}
\usepackage{arydshln}
\newcommand{\vect}[1]{\boldsymbol{\mathbf{#1}}}
\newcommand{\mat}[1]{\boldsymbol{\mathbf{#1}}}

\title{CAPA: Contribution-Aware Pruning and FFN Approximation\\for Efficient Large Vision-Language Models}

\author{
  \textbf{Samyak Jha} \\
  Indian Institute of Technology (ISM) \\
  Dhanbad, India \\
  \texttt{samyakjha71@gmail.com} \\\And
  \textbf{Junho Kim}\thanks{Corresponding author.} \\
  University of Illinois Urbana-Champaign\\
  Urbana, IL, USA \\
  \texttt{arkimjh@illinois.edu}
}

\begin{document}
\maketitle
\begin{abstract}
Efficient inference in Large Vision-Language Models is constrained by the high cost of processing thousands of visual tokens, yet it remains unclear which tokens and computations can be safely removed. While attention scores are commonly used to estimate visual token importance, they are an imperfect proxy for actual contribution. We show that Attention Contribution, which weights attention probabilities by value vector magnitude, provides a more accurate criterion for visual token selection. Our empirical analysis reveals that visual attention sinks are functionally heterogeneous, comprising Probability Dumps with low contribution that can be safely pruned, and Structural Anchors with high contribution essential for maintaining model performance. Further, we identify substantial redundancy in Feed-Forward Networks (FFNs) associated with visual tokens, particularly in intermediate layers where image tokens exhibit linear behavior. Based on our findings, we introduce CAPA (Contribution-Aware Pruning and FFN Approximation), a dual-strategy framework that prunes visual tokens using attention contribution at critical functional transitions and reduces FFN computation through efficient linear approximations. Experiments on various benchmarks across baselines show that CAPA achieves competent efficiency--performance trade-offs with improved robustness.
\end{abstract}

\section{Introduction}
Large Vision Language Models (LVLMs) such as LLaVA \cite{liu2023visual}, Qwen VL \cite{bai2025qwen2,wang2024qwen2}, and InternVL \cite{chen2024expanding,chen2024internvl,zhu2025internvl3} have achieved remarkable success by bridging the gap between visual perception and textual reasoning.

However, the computational cost of processing high resolution images, often involving thousands of visual tokens, remains a major bottleneck. While visual token pruning has emerged as a popular solution, most existing methods \cite{chen2024image,endo2025feather,ye2025fit,zhang2025vispruner} rely heavily on attention scores or heuristic metrics derived directly from the attention scores to identify unimportant tokens.

In addition, there have been several works showing that the language model backbones in LVLMs treat image tokens and text tokens separately~\cite{li2025redundancylens,zhang2024treat,jyoti2025free}, and hence their redundancy should be handled independently. While prior research in the LLM domain has explored approximating the Feed-Forward Networks (FFNs) through methods such as sparse approximation~\cite{frantar2023sparsegpt,zhang2022moefication,liu2023deja,chavan2024surgical}, these techniques have not been extended to LVLMs, specifically regarding vision tokens. This gap is critical because, as we show further, vision tokens and text tokens do not behave in the same way. Consequently, redundancy in FFNs for vision tokens across layers remains largely unexplored, particularly in identifying which layers exhibit FFN redundancy and which do not.

\begin{figure*}[t]
    \centering
    \includegraphics[width=\textwidth]{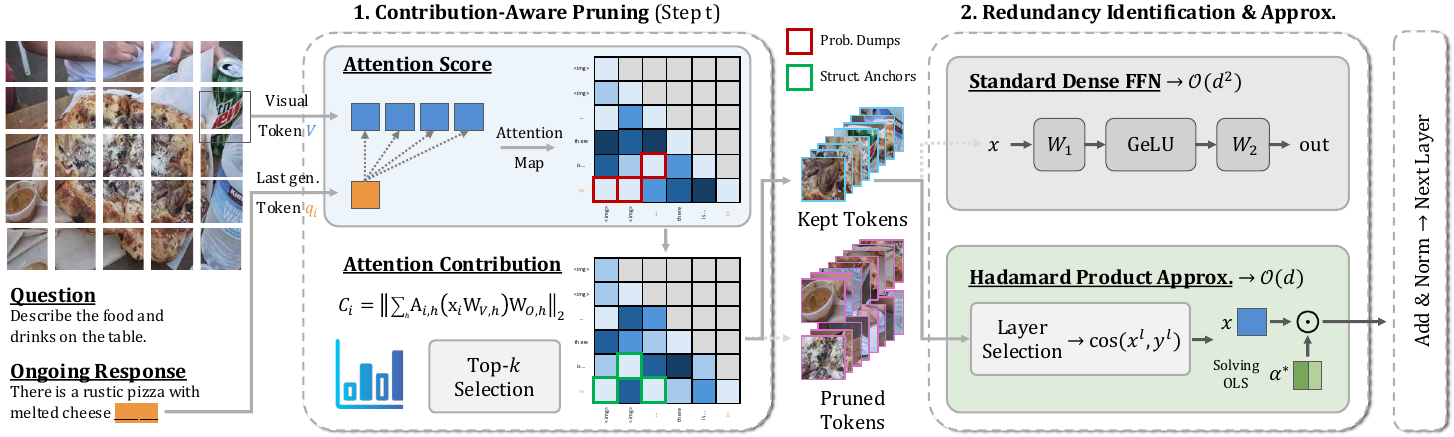}
    \vspace{-3mm}
    \caption{Overall framework of CAPA~\textemdash~(Left) Contribution-Aware Pruning: at each generation step $t$, we compute the \textit{Attention Contribution} score $C_i$ for every visual token by weighting its value vector magnitude with the attention probability assigned by the last generated token query $q_t$;
    (Right) FFN Approximation: in the layers identified as redundant (high input-output cosine similarity), we replace the computationally expensive dense FFNs ($\mathcal{O}(d^2)$) with a lightweight, learned element-wise Hadamard product ($\mathcal{O}(d)$).}
    \label{fig:capa}
\end{figure*}

Through empirical analyses of visual token behaviors in LVLMs, we identify two key observations that existing acceleration methods fail to capture. First, attention scores alone are an insufficient proxy for visual token importance, as they do not account for the magnitude of the associated value vectors, motivating attention contribution as a more faithful signal for visual token pruning. Second, redundancy in LVLMs extends beyond token selection to the FFNs associated with visual tokens, where we observe that, in certain layers, FFN transformations exhibit near-linear behavior.

Building on these observations, we propose CAPA (Contribution-Aware Pruning and FFN Approximation), a dual-strategy framework for efficient LVLM inference. CAPA directly operationalizes our empirical findings by addressing redundancy at both the token and computation levels: it prunes visual tokens based on attention contribution rather than raw attention scores, and reduces unnecessary computation by approximating FFN transformations for vision tokens in layers where redundancy is observed. By jointly considering which visual tokens matter and where visual computation can be reduced, CAPA is designed to achieve favorable efficiency–performance trade-offs without compromising model robustness. Extensive experiments across multiple LVLM backbones and benchmarks show that CAPA consistently preserves task performance while delivering substantial inference acceleration, outperforming existing pruning-based baselines.

Our contributions are as follows:
\begin{enumerate}[topsep=2pt, itemsep=2pt, parsep=0pt]
    \item We present empirical analyses of visual token processing in LVLMs, showing that attention scores alone are insufficient to capture visual token importance and that significant redundancy exists in FFNs associated with vision tokens across layers.
    \item Based on our empirical findings, we propose CAPA, a dual-strategy framework that combines contribution-aware visual token pruning with FFN approximation to reduce inference cost in LVLMs.
    \item Through extensive experiments on multiple LVLM backbones and benchmarks, we demonstrate that CAPA achieves strong efficiency--performance trade-offs while maintaining robust model performance.
\end{enumerate}

\section{Related Work}

\paragraph{Vision Token Pruning in LVLMs.}
Reducing the computational overhead of Large vision–language models has attracted growing attention, with visual token pruning emerging as a dominant acceleration strategy. Early approaches primarily rely on raw attention scores or heuristic importance measures to discard visually redundant tokens \cite{chen2024image,endo2025feather}. These methods implicitly assume that attention probability alone reflects token saliency. However, recent analyses have shown that attention weights can be misleading when detached from value representations, particularly in multi-modal settings \cite{kobayashi2020attention,basu2024understanding,guo2024attention}. In contrast to prior work, we explicitly incorporate value vector magnitude through Attention Contribution, yielding a more faithful estimate of a token’s functional impact on downstream representations and enabling more reliable visual token pruning.

\begin{figure*}[t]
    \centering
    \includegraphics[width=0.99\textwidth]{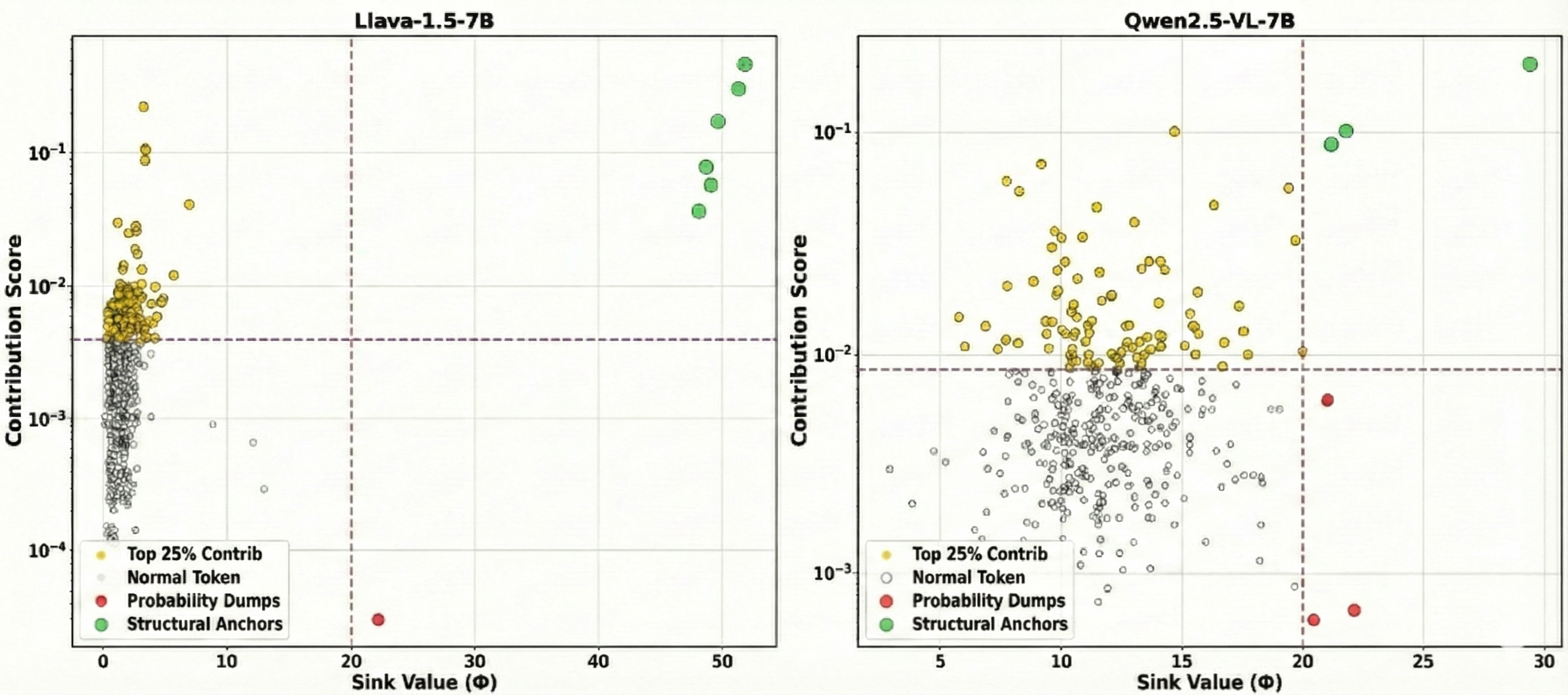}
    \caption{Sink distribution analysis across LLaVA-1.5 and Qwen2.5-VL. Using a sink threshold of $\tau=20$, we classify tokens into two groups: low-contribution \textbf{Probability Dumps} (Type I) and high-contribution \textbf{Structural Anchors} (Type II) . The dashed line represents the sink identification threshold.}
    \label{fig:sink_taxonomy}
\end{figure*}

\paragraph{Visual Attention Sinks.}
Several studies have observed the emergence of attention sinks particularly visual attention sinks in LVLMs, where certain tokens attract disproportionately high attention mass \cite{kang2025see,queipo2025attention,xiao2023efficient}. These works typically treat attention sinks as a single phenomenon and focus on mitigating their negative effects through masking or reweighting. Our work departs from this view by demonstrating that visual attention sinks are functionally heterogeneous. Using attention contribution, we distinguish between low-contribution probability dumps and high-contribution structural anchors, enabling safe pruning without compromising model integrity.
\paragraph{Feed Forward Network Approx. and Efficiency.}
The FFNs constitute approximately two-thirds of the parameter count in Transformer models, making them a prime target for optimization. In the realm of LLMs, extensive research has attempted to address FFN redundancy by replacing standard dense layers with significantly more complex formulations. These include substituting FFNs with Mixture-of-Experts (MoE) architectures to route tokens to specialized expert blocks \cite{zhang2022moefication,gale2211megablocks}, replacing dense matrices with structured linear parameterizations or Monarch matrices \cite{fu2023monarch, wei2024building}. Other works have explored even more radical structural changes, such as approximating FFNs with Kolmogorov-Arnold Networks (KAN) using learnable spline functions \cite{liu2024kan}, deploying spiking neural networks for event-driven sparsity \cite{zhu2023spikegpt}, or applying tensor decomposition like Tucker or CP decomposition to compress weight matrices \cite{wang2403svd, xu2023tensorgpt}. 

To the best of our knowledge no comparable work has investigated FFN redundancy specifically for vision tokens in LVLMs. We bridge this gap by identifying that visual tokens exhibit strong linearity in intermediate layers distinct from text tokens allowing us to replace expensive FFNs with a lightweight, learned Hadamard product rather than complex architectural substitutes.

\section{Empirical Analysis}

\subsection{Deconstructing Visual Attention Sinks}
To revisit the functional role of high-attention visual tokens, we first examine the extent to which statistical presence (high attention weights) aligns with representational impact (contribution to the residual stream). While prior work~\cite{kang2025see} identifies \textit{visual attention sinks} solely based on massive activation in specific hidden dimensions (defined as Sink Value $\phi > \tau$ with $\tau=20$), we observe that this definition may conflate functionally distinct token behaviors. Accordingly, we analyze the token distribution of LLaVA-1.5~\cite{liu2023visual} and Qwen2.5-VL~\cite{bai2025qwen2} by correlating two orthogonal metrics for every visual token: the Sink Value ($\phi$), representing the activation magnitude in outlier dimensions, and the Attention Contribution ($C_i$), representing the weighted value vector's magnitude added to the residual stream (defined in Sec~\ref{sec:contribution}).

As visualized in Fig. \ref{fig:sink_taxonomy}, this multi-dimensional analysis reveals a critical functional dichotomy among tokens that arguably qualify as \textit{sinks} under the standard definition ($\phi > 20$). We observe that visual tokens do not form a monolithic group; rather, they bifurcate into two distinct clusters based on their contribution scores:

\begin{itemize}[left=0pt, noitemsep, topsep=0pt]
    \item \textbf{Type I: Probability Dumps} are
    clustered in the low-contribution region (highlighted in red), these tokens exhibit the classic behavior described in recent literature. Despite possessing high attention weights and high sink values ($\phi > 20$), their resulting contribution $C_i$ is negligible. This confirms that their primary function is to serve as passive receptacles for excess probability mass generated by the Softmax operation, rather than encoding semantic content. Consequently, they represent true redundancy and can be pruned safely.
    \item \textbf{Type II: Structural Anchors} are, crucially, identified as a second cluster (highlighted in green) that standard sink thresholds only fail to distinguish. They couple high sink values with massive value vector magnitudes, resulting in high Attention Contribution ($C_i$). Unlike probability dumps, these tokens act as critical biases or \textit{anchors} within the residual stream. Our analysis shows that while they statistically resemble sinks due to high $\phi$, their removal leads to immediate representational collapse.
\end{itemize}

This empirical distinction demonstrates that attention scores alone are insufficient for pruning while maintaining visual perception capabilities.

\begin{figure}[t]
\centering
\includegraphics[width=1.0\linewidth]{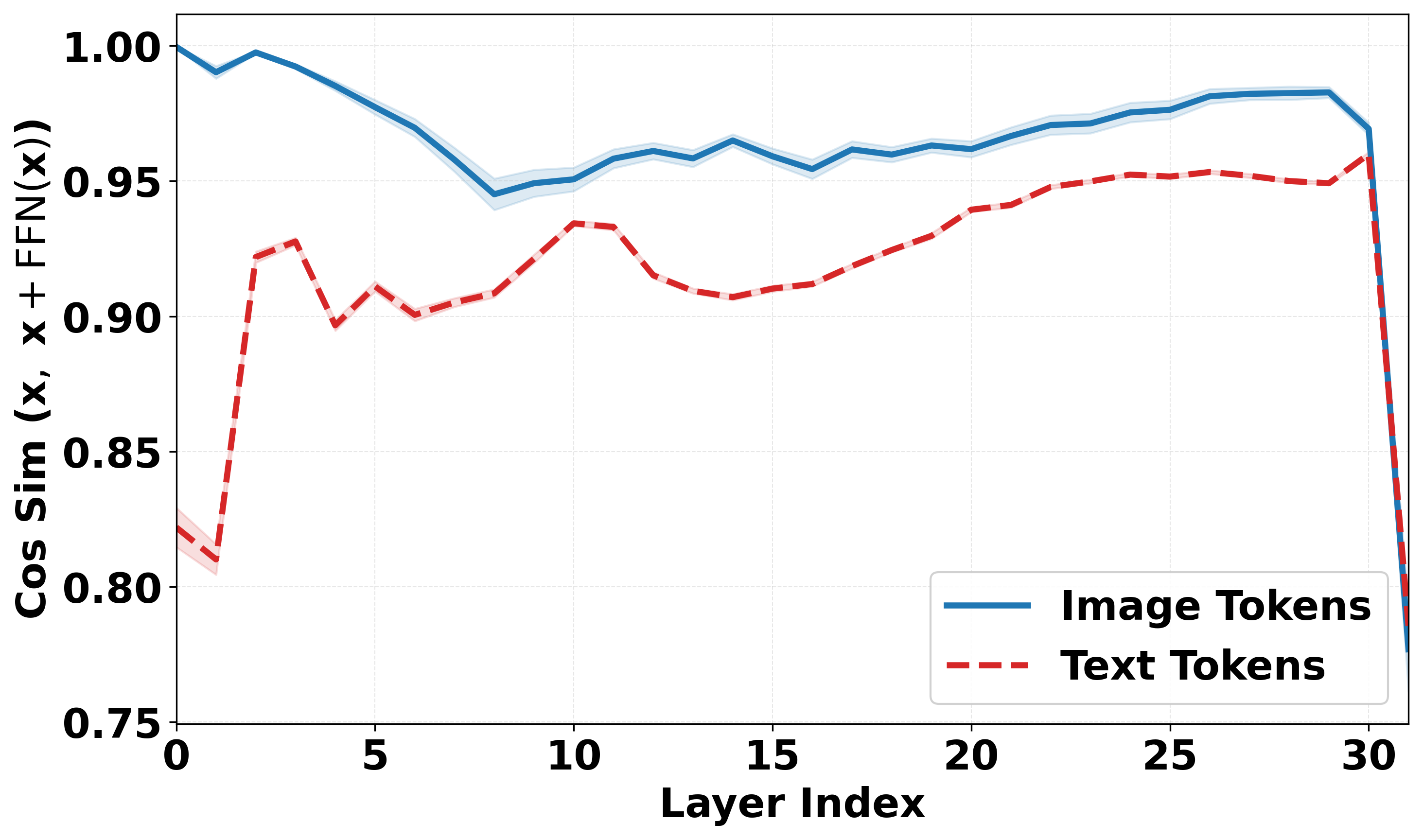}
\vspace{-6mm}
\caption{Layer-wise FFN Redundancy for Image vs. Text Tokens. The plot displays the mean cosine sim. $\cos(\mathbf{x}, \mathbf{x} + \mathrm{FFN}(\mathbf{x}))$ across network layers, evaluated on 500 samples from the MSCOCO 2017 val set.}
\vspace{-4mm}
\label{fig:ffn_cosine}
\end{figure}

\subsection{Modality-Dependent FFN Redundancy}
After re-examining the functional role of visual tokens at the attention level, we now turn to their role in FFNs, which constitute massive parameter count in Transformers and represent a primary computational bottleneck during inference. While prior research in LLMs~\cite{pessoa2023one,gromov2024unreasonable,bercovich2025ffn} has explored FFN redundancy, these works typically treat all tokens uniformly. In contrast, we hypothesize that in LVLMs, the language backbone processes visual and text tokens with different degrees of non-linearity. To delve into this, we design an experiment to quantify the \textit{functional necessity} of FFN layers specifically for visual tokens.

We analyze the linearity of FFN transformations by measuring the directional alignment between the layer input $\mathbf{x}$ and the post-FFN residual output $\mathbf{y} = \mathbf{x} + \text{FFN}(\mathbf{x})$. Intuitively, if an FFN contributes significant non-linear processing, the output vector should diverge effectively from the residual connection. Conversely, a cosine similarity near 1.0 implies an identity-like mapping where the FFN is redundant. We formalize this metric as:
\begin{equation}
    \text{Sim}(\mathbf{x}) = \cos(\mathbf{x}, \mathbf{x} + \text{FFN}(\mathbf{x})).
\end{equation}

We evaluated this metric across all layers of LLaVA-1.5-7B, Qwen2.5-VL-7B, and InternVL3-8B using 500 samples from the MSCOCO 2017 validation set \cite{lin2014microsoft}.

\begin{figure}[t]
\centering
\includegraphics[width=1.0\linewidth]{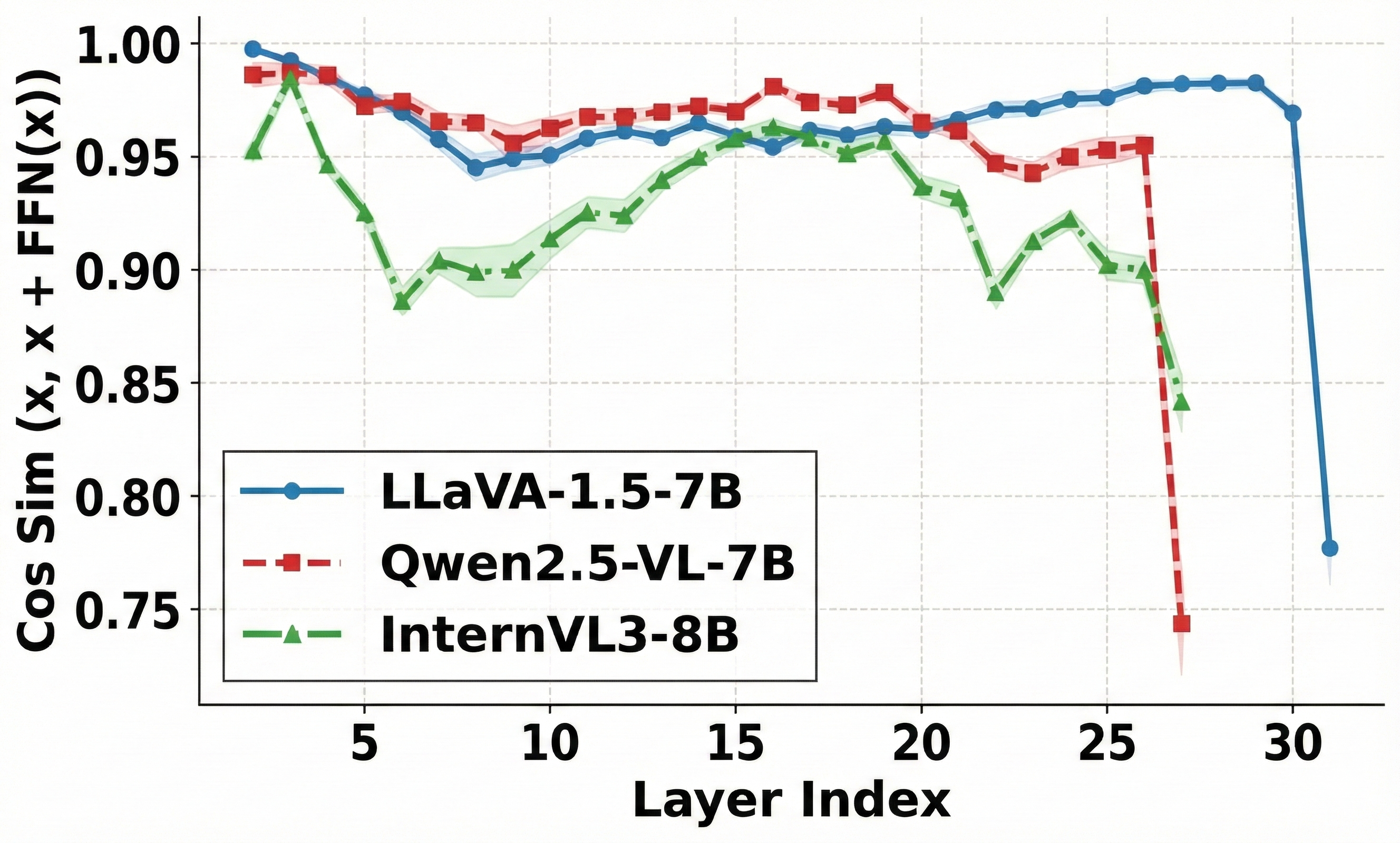}
\vspace{-6mm}
\caption{Comparative analysis of FFN redundancy across LVLM architectures. We plot the average cosine sim. between the input hidden state $\mathbf{x}$ and the residual output $\mathbf{x} + \text{FFN}(\mathbf{x})$ for three baselines.}
\vspace{-3mm}
\label{fig:models_ffn_redundancy}
\end{figure}

Our layer-wise analysis, as visualized in Fig. \ref{fig:ffn_cosine}, reveals a striking disparity in processing patterns between modalities. Image tokens (solid blue line) consistently maintain near-identity similarity ($>0.96$) across the majority of layers. In contrast, text tokens (dashed red line) exhibit lower similarity in early and middle layers, indicating a higher reliance on FFN transformations, before converging to high redundancy in the deeper layers. Both modalities show a significant drop in similarity at the final layer. This distinction is not unique to a single architecture; as demonstrated in Fig. \ref{fig:models_ffn_redundancy}, this pattern of high visual token linearity in intermediate layers is consistent across three baselines.

This structural redundancy suggests that for visual tokens, the expensive FFN computations in these layers effectively collapse into linear transformations. This indicates that the model's depth is not fully utilized for visual processing, motivating our proposed strategy to replace these dense layers with efficient linear approximations without compromising representational integrity.

\section{Methodology}
Taken together, our empirical analyses show that inefficiency in LVLM inference arises from complementary sources. At the attention level, visual tokens with high attention mass exhibit heterogeneous functional roles, rendering attention score pruning unreliable. At the computation level, FFN transformations for visual tokens exhibit substantial linear redundancy. These findings suggest that effective acceleration requires jointly reasoning about which visual tokens matter and where visual computation can be simplified, motivating CAPA, a dual-strategy framework that integrates contribution-aware visual token pruning with FFN approximation as summarized in~Fig.~\ref{fig:capa}.

\subsection{Contribution-Aware Pruning}
\label{sec:contribution}
To effectively identify redundant visual tokens, it is crucial to distinguish between the probability of attending to a token and the actual informational impact of that token on the network's processing. Standard pruning methods have solely exploited attention scores, implicitly assuming that higher attention weights correlate directly with feature importance. However, attention weights function primarily as routing coefficients—they determine which tokens are selected, but not necessarily the magnitude of the update applied to the residual stream. The actual quantity of information transferred depends on the Value vectors projected by the attention heads. A token with a high attention score but a negligible value vector magnitude contributes little to the representation.

To address it, we quantify the importance of a visual token $i$ using its attention contribution score $C_i$, which integrates both the routing weight and the feature magnitude. Specifically, we compute the weighted aggregation of the value vectors projected by the output matrix $\mat{W}_{O,h}$ to capture the total magnitude of information transfer. It is defined as the $\ell_2$ norm of this contribution to the residual stream:
\begin{equation}
    C_i = \left\| \sum_{h=1}^{H} \mat{A}_{i,h} \bigl(\vect{x}_i \mat{W}_{V,h}\bigr)\mat{W}_{O,h} \right\|_2,
\end{equation}
where $\mat{A}_{i,h}$ denotes the attention probability assigned to token $i$ by head $h$, $\vect{x}_i$ is the input visual token representation, and $\mat{W}_{V,h}$, $\mat{W}_{O,h}$ are the value and output projection matrices. By measuring the norm of the projected vector, this formulation captures the effective contribution of each token to the residual update, providing a more faithful proxy for token saliency than raw attention weights alone.

In practice, we apply this metric dynamically during the generation process. At every generation step, we calculate $C_i$ relative to the current query token (the last generated token). We then select the top-$k$ visual tokens with the highest contribution scores to be retained in the key-value cache, while tokens with minimal contribution are pruned. This ensures that the retained visual context is not static; instead, it dynamically adapts to the specific semantic requirements of the current generation step, preserving only the visual information that functionally alters the model's output distribution.

\subsection{Redundancy Identification \& Approx.}
\label{sec:method_approx}

Having addressed the redundancy in the sequence length ($N$) through token pruning, we essentially employ a dual-strategy approach by next targeting the redundancy in the model's width $d$. As in Fig. \ref{fig:capa}, we delve into FFNs, which account for the majority of the remaining FLOPs, to estimate their necessity and propose a lightweight approximation.

\paragraph{Layer Selection Strategy.}
To identify layers suitable for approximation, we analyze the transformation magnitude of the FFNs across the model. We define the redundancy of a layer $l$ by the cosine similarity between its input hidden state $\mathbf{x}^{(l)}$ and the output of the residual block $\mathbf{y}^{(l)} = \mathbf{x}^{(l)} + \text{FFN}(\mathbf{x}^{(l)})$. As we observed in Fig.~\ref{fig:models_ffn_redundancy}, layers exhibiting high cosine similarity imply a near-identity transformation, where the FFN contributes minimal non-linear modification to the feature space.

\begin{table*}[t]
\caption{Pruning performance comparison of CAPA against baselines \textit{L}: LLaVA-1.5-7B, \textit{Q}: Qwen2.5-VL-7B, and \textit{I}: InternVL3-8B. CAPA demonstrates consistent robustness and superior performance across all benchmarks.}
\vspace{-3mm}
\label{table:main}
\centering
\small
\setlength{\tabcolsep}{2.0pt}
\resizebox{1.0\linewidth}{!}{
\begin{tabular}{cl ccc ccc ccc ccc ccc ccc}
\Xhline{2\arrayrulewidth}
\multirow{2}{*}[-0.7ex]{\textbf{$l$}} & \multirow{2}{*}[-0.7ex]{\textbf{Method}} & \multicolumn{3}{c}{\textbf{VQA2}} & \multicolumn{3}{c}{\textbf{MMBench}} & \multicolumn{3}{c}{\textbf{MMVet}} & \multicolumn{3}{c}{\textbf{TextVQA}} & \multicolumn{3}{c}{\textbf{SEED}} & \multicolumn{3}{c}{\textbf{MMMU}} \\
\cmidrule(lr){3-5} \cmidrule(lr){6-8} \cmidrule(lr){9-11} \cmidrule(lr){12-14} \cmidrule(lr){15-17} \cmidrule(lr){18-20}
& & \textit{L} & \textit{Q} & \textit{I} & \textit{L} & \textit{Q} & \textit{I} & \textit{L} & \textit{Q} & \textit{I} & \textit{L} & \textit{Q} & \textit{I} & \textit{L} & \textit{Q} & \textit{I} & \textit{L} & \textit{Q} & \textit{I} \\ 
\specialrule{\lightrulewidth}{0pt}{0pt}

\specialrule{\lightrulewidth}{0pt}{0pt}
\rowcolor[gray]{0.95} Full & Vanilla & 76.55 & 83.74 & 80.28 & 62.97 & 82.21 & 84.71 & 28.89 & 61.83 & 70.41 & 48.64 & 83.69 & 80.89 & 60.14 & 76.80 & 74.97 & 34.67 & 48.68 & 55.44 \\ 
\specialrule{\lightrulewidth}{0pt}{0.5ex}
& Unif. & 72.79 & 82.61 & 79.56 & 61.25 & 74.10 & 82.65 & 24.03 & 59.31 & 67.56 & 32.98 & 80.82 & 81.34 & 56.87 & 71.20 & 73.92 & 34.62 & 47.10 & 54.67 \\
& FastV   & 72.14 & 82.73 & 78.26 & 62.37 & 75.34 & 82.71 & 24.67 & 60.91 & 68.80 & 42.56 & 82.24 & 78.87 & 56.01 & 73.80 & 72.02 & 34.56 & 47.67 & 53.42 \\
& Feather & 73.87 & 82.71 & 79.26 & 62.45 & 75.10 & 82.70 & 26.30 & 60.68 & 69.95 & 45.30 & 81.24 & 79.87 & 56.86 & 73.10 & 73.02 & 34.70 & 47.67 & 54.43 \\ 
\cdashline{3-20}[0.5pt/2pt]
\noalign{\vskip 0.1ex}
\rowcolor[HTML]{ECF4FF} \cellcolor{white} \multirow{-4}{*}{\rotatebox[origin=c]{90}{Early}} & \textbf{CAPA} & \textbf{74.76} & \textbf{83.10} & \textbf{80.26} & \textbf{62.80} & \textbf{78.16} & \textbf{84.79} & \textbf{28.75} & \textbf{62.51} & \textbf{70.37} & \textbf{46.70} & \textbf{82.90} & \textbf{80.92} & \textbf{57.91} & \textbf{75.30} & \textbf{74.02} & \textbf{35.44} & \textbf{47.97} & \textbf{55.44} \\ 
\specialrule{\lightrulewidth}{0pt}{0.5ex}
& Unif. & 73.95 & 82.61 & 79.05 & 62.60 & 75.20 & 84.62 & 25.69 & 62.11 & 68.34 & 36.07 & 81.74 & 80.65 & 58.80 & 73.92 & 73.00 & 35.44 & 47.67 & 55.00 \\
& FastV   & 75.27 & 82.75 & 79.50 & 62.43 & 75.93 & 83.10 & 27.01 & 61.42 & 69.31 & 47.05 & 81.55 & 79.10 & 56.91 & 73.65 & 71.27 & 35.56 & 47.69 & 54.27 \\
& Feather & 75.28 & 82.79 & 79.56 & 62.54 & 75.30 & 83.20 & 26.91 & 61.83 & 69.95 & 47.46 & 82.10 & 80.12 & 57.05 & 73.69 & 72.02 & 35.67 & 47.67 & 53.21 \\ 
\cdashline{3-20}[0.5pt/2pt]
\noalign{\vskip 0.1ex}
\rowcolor[HTML]{ECF4FF} \cellcolor{white} \multirow{-4}{*}{\rotatebox[origin=c]{90}{Transition}} & \textbf{CAPA} & \textbf{76.21} & \textbf{84.10} & \textbf{80.26} & \textbf{62.82} & \textbf{77.20} & \textbf{84.79} & \textbf{27.96} & \textbf{64.25} & \textbf{69.85} & \textbf{48.40} & \textbf{82.34} & \textbf{81.20} & \textbf{59.94} & \textbf{75.72} & \textbf{74.02} & \textbf{36.61} & \textbf{47.89} & \textbf{55.44} \\ 
\specialrule{\lightrulewidth}{0pt}{0.5ex}
& Unif. & 74.65 & 82.62 & 79.90 & 62.47 & 75.10 & 83.00 & 25.55 & 59.95 & 69.49 & 37.87 & 81.78 & 80.35 & 59.17 & 71.70 & 73.02 & 34.22 & 48.44 & 51.40 \\
& FastV   & 75.27 & 82.75 & 80.10 & 62.43 & 76.40 & 84.30 & 28.27 & 61.74 & 69.27 & 47.38 & 82.37 & 80.98 & 59.13 & 73.75 & 72.50 & 35.67 & 48.54 & 51.60 \\
& Feather & 75.52 & 82.80 & 80.30 & 62.43 & 76.70 & 84.70 & 28.48 & 63.30 & 69.35 & 47.43 & 82.39 & 80.98 & 59.13 & 73.79 & 71.79 & 35.68 & 48.59 & 51.67 \\ 
\cdashline{3-20}[0.5pt/2pt]
\noalign{\vskip 0.1ex}
\rowcolor[HTML]{ECF4FF} \cellcolor{white} \multirow{-4}{*}{\rotatebox[origin=c]{90}{Late}} & \textbf{CAPA} & \textbf{76.20} & \textbf{84.30} & \textbf{80.27} & \textbf{62.67} & \textbf{77.70} & \textbf{84.79} & \textbf{28.70} & \textbf{64.70} & \textbf{69.63} & \textbf{48.38} & \textbf{82.30} & \textbf{81.93} & \textbf{60.08} & \textbf{75.60} & \textbf{72.02} & \textbf{35.89} & \textbf{49.33} & \textbf{53.67} \\ 
\Xhline{2\arrayrulewidth}
\end{tabular}
}
\vspace{-3mm}
\end{table*}

We formalize our selection criterion $\mathcal{S}$ with a threshold $\eta$. A layer is selected for approximation if the similarity over the calibration set exceeds $\eta$:
\begin{equation}
    \mathcal{S} = \{ l \mid \mathbb{E}_{\mathbf{x} \sim \mathcal{D}} \left[ \text{cos}(\mathbf{x}^{(l)}, \mathbf{y}^{(l)}) \right] > \eta \}
\end{equation}
Based on this analysis, we target layers that fall within high-redundancy regions while preserving critical processing stages (\textit{e.g.,} final output layers). 

\paragraph{Optimized Hadamard Product Approximation.}
For layers $l \in \mathcal{S}$, we replace the computationally expensive matrix operations of the FFN with a learned, lightweight element-wise scaling vector $\boldsymbol{\alpha} \in \mathbb{R}^{d}$. This collapses the entire residual FFN block into a single Hadamard product operation:
\begin{equation}
    \mathbf{y}^{(l)} \approx \hat{\mathbf{y}}^{(l)} = \mathbf{x}^{(l)} \odot \boldsymbol{\alpha}
\end{equation}
where $\odot$ denotes the element-wise product. It reduces the layer's FLOPs count from $\mathcal{O}(d^2)$ to $\mathcal{O}(d)$. More discussion on the complexity in~\ref{sec:appendix_complexity}

\paragraph{Closed-Form Least Squares Solution.}
To find the optimal $\boldsymbol{\alpha}$, we avoid iterative gradient descent and instead solve the Ordinary Least Squares (OLS) objective analytically. We collect activation statistics using 500 samples from the COCO 2017 train set~\cite{lin2014microsoft}. For each feature dimension $k{\in}\{1, \dots, d\}$, we minimize the squared reconstruction error between the approximated output and the true residual output over $N$ calibration samples:
\begin{equation}
    \min_{\alpha_k} \sum_{n=1}^{N} \left( \alpha_k x_{n,k} - y_{n,k} \right)^2.
\end{equation}

Setting the partial derivative with respect to $\alpha_k$ to zero yields a closed-form solution based on the second-order moments of the features:
\begin{equation}
    \alpha_k^* = \frac{\sum_{n=1}^{N} x_{n,k} \cdot y_{n,k}}{\sum_{n=1}^{N} x_{n,k}^2}.
\end{equation}
In vector notation, this is computed as $\boldsymbol{\alpha}^* = (\sum \mathbf{x}_n \odot \mathbf{y}_n) \oslash (\sum \mathbf{x}_n \odot \mathbf{x}_n)$, where $\oslash$ represents element-wise division. This approach ensures $\boldsymbol{\alpha}$ captures the optimal linear scaling factor that statistically minimizes error across the data distribution. We observe that feature statistics converge rapidly, making 500 samples sufficient for a robust approximation without extensive retraining.

\section{Experiments}
\label{sec:experiments}

\subsection{Experimental Setup}
\paragraph{Baseline Models and Benchmarks.}
We evaluate our framework on three state-of-the-art LVLMs: LLaVA-1.5-7B \cite{liu2023visual}, Qwen2.5-VL-7B \cite{bai2025qwen2}, and InternVL3-8B \cite{zhu2025internvl3}. To ensure a comprehensive assessment of multi-modal capabilities, we report results on six standard benchmarks: VQAv2 \cite{goyal2017making} for general visual QA, MMBench \cite{liu2024mmbench} and SEED-Bench \cite{li2023seed} for holistic perception, MM-Vet \cite{yu2023mm} for integrated reasoning, TextVQA \cite{singh2019towards} for OCR capabilities, and MMMU \cite{yue2024mmmu} for multi-discipline expert reasoning.

\begin{figure*}[t]
    \centering
    \includegraphics[width=\linewidth]{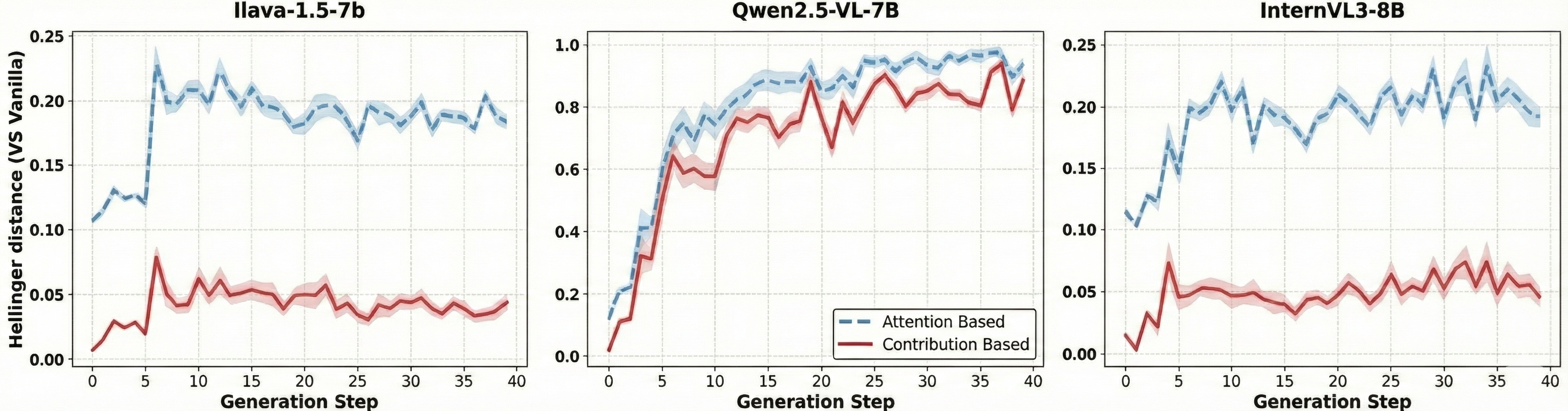}
    \vspace{-8mm}
    \caption{Hellinger distance between the vanilla model and pruned logits across generation steps, comparing attention-score and attention-contribution pruning for 3 baselines on the COCO2017 val. set (200 examples).}
    \label{fig:logit_divergence}
\end{figure*}

\begin{table*}[t!]
\caption{Impact of different FFN replacement strategies applied to both the Vanilla (unpruned) and CAPA (pruned) models. The Hadamard product recovers most of the performance degradation caused by skipping FFNs. \textit{L}: LLaVA-1.5-7B, \textit{Q}: Qwen2.5-VL-7B, and \textit{I}: InternVL3-8B.}
\label{tab:ffn_results}
\vspace{-3mm}
\centering
\small
\setlength{\tabcolsep}{2.0pt}
\resizebox{1.0\linewidth}{!}{
\begin{tabular}{cl ccc ccc ccc ccc ccc ccc}
\Xhline{2\arrayrulewidth}
\multirow{2}{*}[-0.7ex]{\textbf{Type}} & \multirow{2}{*}[-0.7ex]{\textbf{Strategy}} & \multicolumn{3}{c}{\textbf{VQA2}} & \multicolumn{3}{c}{\textbf{MMBench}} & \multicolumn{3}{c}{\textbf{MMVet}} & \multicolumn{3}{c}{\textbf{TextVQA}} & \multicolumn{3}{c}{\textbf{SEED}} & \multicolumn{3}{c}{\textbf{MMMU}} \\
\cmidrule(lr){3-5} \cmidrule(lr){6-8} \cmidrule(lr){9-11} \cmidrule(lr){12-14} \cmidrule(lr){15-17} \cmidrule(lr){18-20}
& & \textit{L} & \textit{Q} & \textit{I} & \textit{L} & \textit{Q} & \textit{I} & \textit{L} & \textit{Q} & \textit{I} & \textit{L} & \textit{Q} & \textit{I} & \textit{L} & \textit{Q} & \textit{I} & \textit{L} & \textit{Q} & \textit{I} \\ 
\specialrule{\lightrulewidth}{0pt}{0pt}

\specialrule{\lightrulewidth}{0pt}{0.5ex}
& Full Model & 76.55 & 83.74 & 80.28 & 62.97 & 82.21 & 84.71 & 28.89 & 61.83 & 70.41 & 48.64 & 83.69 & 80.89 & 60.14 & 76.80 & 74.97 & 34.67 & 48.68 & 55.44 \\ 
& Skip FFN   & 74.02 & 79.15 & 75.70 & 58.12 & 72.40 & 78.13 & 22.45 & 49.30 & 59.10 & 44.10 & 76.12 & 74.90 & 56.78 & 69.20 & 67.32 & 29.30 & 38.50 & 42.37 \\ 
\cdashline{3-20}[0.5pt/2pt]
\noalign{\vskip 0.1ex}
\rowcolor[HTML]{ECF4FF} \cellcolor{white} \multirow{-3}{*}{\rotatebox[origin=c]{90}{Vanilla}} & \textbf{Hadamard} & \textbf{75.48} & \textbf{81.02} & \textbf{78.79} & \textbf{61.32} & \textbf{77.85} & \textbf{80.52} & \textbf{26.90} & \textbf{56.40} & \textbf{65.05} & \textbf{47.15} & \textbf{80.22} & \textbf{78.03} & \textbf{58.82} & \textbf{73.15} & \textbf{71.95} & \textbf{32.85} & \textbf{44.10} & \textbf{49.33} \\ 

\specialrule{\lightrulewidth}{0pt}{0.5ex}
& Prune Only & 76.20 & 84.30 & 80.27 & 62.67 & 77.70 & 84.79 & 28.70 & 64.70 & 69.63 & 48.38 & 82.30 & 81.93 & 60.08 & 75.60 & 72.02 & 35.89 & 49.33 & 53.67 \\ 
& Skip FFN   & 73.88 & 79.60 & 73.90 & 57.90 & 67.20 & 75.80 & 22.10 & 52.15 & 56.45 & 43.85 & 74.50 & 73.12 & 56.40 & 67.90 & 64.50 & 30.25 & 39.45 & 40.85 \\ 
\cdashline{3-20}[0.5pt/2pt]
\noalign{\vskip 0.1ex}
\rowcolor[HTML]{ECF4FF} \cellcolor{white} \multirow{-3}{*}{\rotatebox[origin=c]{90}{CAPA}} & \textbf{Hadamard} & \textbf{75.10} & \textbf{81.85} & \textbf{76.70} & \textbf{61.05} & \textbf{73.20} & \textbf{78.34} & \textbf{26.44} & \textbf{59.12} & \textbf{62.30} & \textbf{46.90} & \textbf{78.90} & \textbf{75.08} & \textbf{58.65} & \textbf{71.42} & \textbf{71.30} & \textbf{34.02} & \textbf{45.22} & \textbf{45.33} \\ 

\Xhline{2\arrayrulewidth}
\end{tabular}
}
\vspace{-3mm}
\end{table*}

\paragraph{Pruning Baselines.}
We compare CAPA against a various set of pruning strategies to validate its efficacy. As foundational baselines, we include the \textbf{Vanilla} (unpruned) model and a \textbf{Uniform} baseline, which naively samples tokens with a fixed stride to match the target token budget. Among attention-centric methods, we evaluate \textbf{FastV} \cite{chen2024image}, a training-free approach that prunes tokens based on raw attention scores in early layers, and \textbf{Feather} \cite{endo2025feather}, which modifies causal masking constraints. For Feather, we implement the method without its ensemble mechanism to isolate the efficacy of its \textit{No-RoPE} heuristic for fair comparison.

\paragraph{Implementation Details.}
For all pruning experiments, we retain 25\% of the visual tokens. Following prior observations~\cite{luo2025direct, yu2025multimodal} that multi-modal transformers exhibit depth-dependent behavior, we partition layers into three stages: \textit{early}, \textit{transition}, and \textit{late}. Pruning is applied at the phase-transition layers identified in \cite{yu2025multimodal} (LLaVA: Layers 5, 12, 16; QwenVL: Layers 3, 11, 21; InternVL: Layers 3, 11, 21). For FFN approximation, we target the high-redundancy blocks as highlighted in Sec. \ref{sec:method_approx} and we keep $\eta$ as 0.96 and hence find the redundant layers (LLaVA: Layers 2--5 \& 22--29; Qwen: Layers 2--5 \& 13--19 ; InternVL Layers 2--4 \& 14--20). The approximation parameters $\alpha$ are calibrated using 500 randomly sampled images from the COCO train set~\cite{lin2014microsoft}.

\subsection{Main Results: Pruning Performance}
Tab.~\ref{table:main} reports layer-wise pruning results for three baselines, with evaluations conducted at critical \textit{phase transition} layers to assess robustness under representational shifts. Across all models, pruning robustness consistently improves in later layers, indicating that dependence on dense visual tokens diminishes as representations mature, rendering image tokens increasingly redundant in deeper layers.

\begin{figure}[t]
\centering
\includegraphics[width=0.99\linewidth]{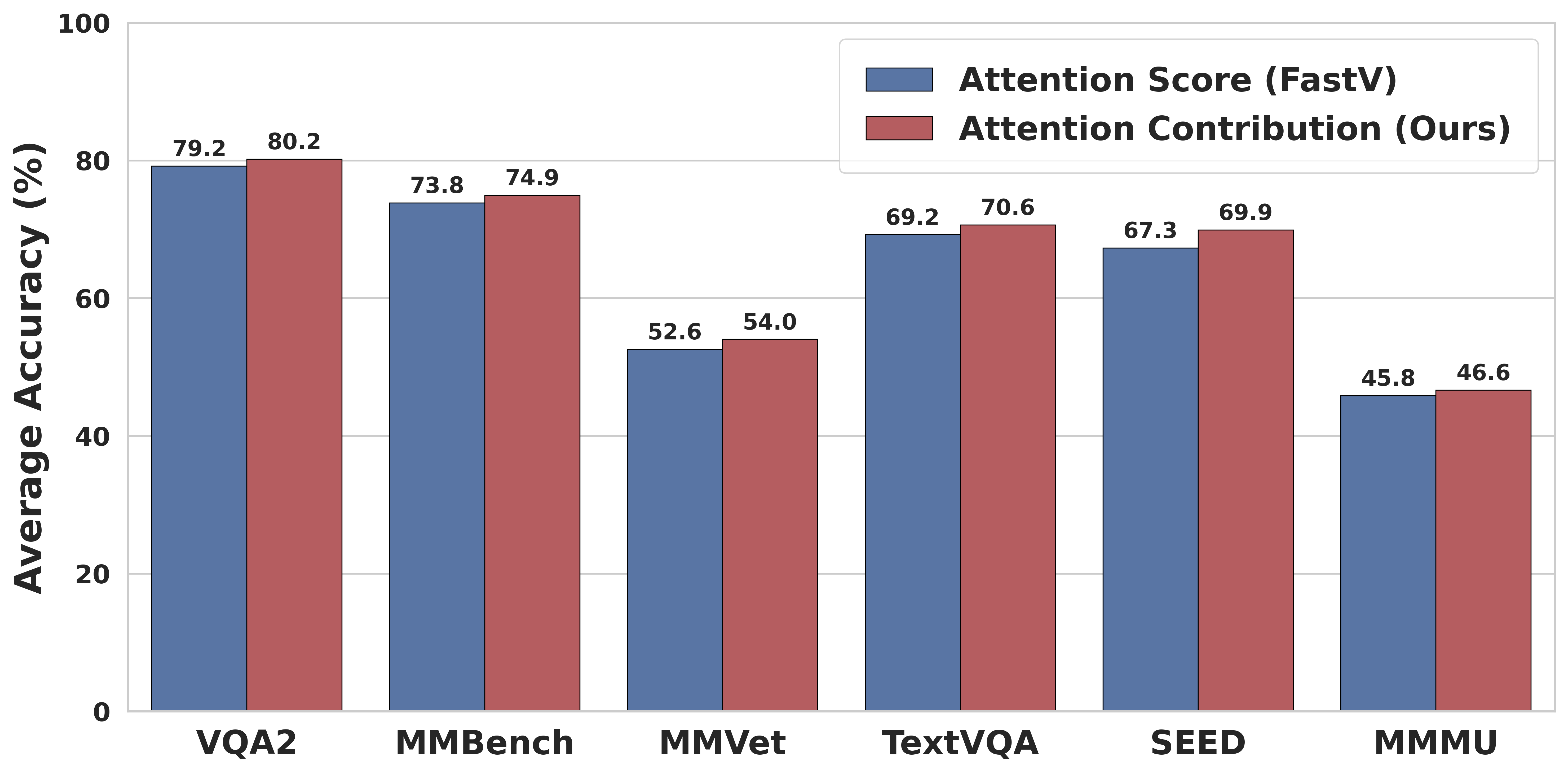}
\vspace{-8mm}
\caption{Performance comparison of standard Attention Score pruning (FastV) against our Attention Contribution method (CAPA) at the transition layers. Results are averaged across 3 baselines for generalizability.}
\vspace{-3mm}
\label{fig:transition}
\end{figure}

CAPA demonstrates a clear advantage at the \textit{transition} layers (Fig.~\ref{fig:transition}), which represent the most sensitive stage for token reduction. Baseline methods such as FastV and Feather show substantial performance degradation in this phase, whereas CAPA maintains performance close to the unpruned Vanilla model. Task-level analysis further shows that CAPA is particularly effective on benchmarks requiring fine-grained perception and complex reasoning (TextVQA, MMMU). This suggests that semantically dense tokens, such as image-embedded text or intricate visual details, often exhibit high value vector magnitudes despite variable attention scores. By explicitly weighting these contributions, CAPA avoids removing information-dense tokens, a common failure mode of attention-only or uniform pruning. Finally, pruning in \textit{late} layers is universally safe, with all methods converging to near-Vanilla performance, validating the high redundancy of visual tokens in the final layers.

\subsection{Ablation Study for CAPA}
\paragraph{Logit Divergence under Attention Pruning.}
To compare how different attention-based pruning criteria preserve the generative behavior of the vanilla model, we measure the divergence between vanilla models and their pruned variants during long-form generation. All tokens before the pruning step are kept identical, ensuring that any deviation is solely induced by the choice of pruning criterion.

Specifically, let $x$ denote the visual input, $\tilde{x}$ the pruned visual input, and $y_{<t}$ the generated prefix up to step $t$. We compare the output distributions
$p(\cdot \mid y_{<t}, x)$ from the vanilla model and $\tilde{p}(\cdot \mid y_{<t}, \tilde{x})$ from the pruned model, where pruning is performed using either attention scores or attention contribution.

We quantify the discrepancy between these distributions using Hellinger distance~\cite{lecam2012asymptotic}:
\begin{equation}
H\big(p,\tilde{p}\big)
=\frac{1}{\sqrt{2}}
\left(\sum_i\left(\sqrt{p_i}-\sqrt{\tilde{p}_i}\right)^2\right)^{1/2},
\end{equation}
where $p_i$ and $\tilde{p}_i$ denote the probabilities assigned to token $i$ by vanilla and pruned models, respectively.

As shown in Fig.~\ref{fig:logit_divergence}, pruning based on attention contribution consistently yields lower Hellinger distances than pruning based on raw attention scores across all evaluated models. This gap widens at later generation steps, indicating that attention contribution more faithfully preserves the vanilla model’s output distribution over long-horizon multi-modal generation.

\begin{figure}[t]
\centering
\includegraphics[width=0.95\linewidth]{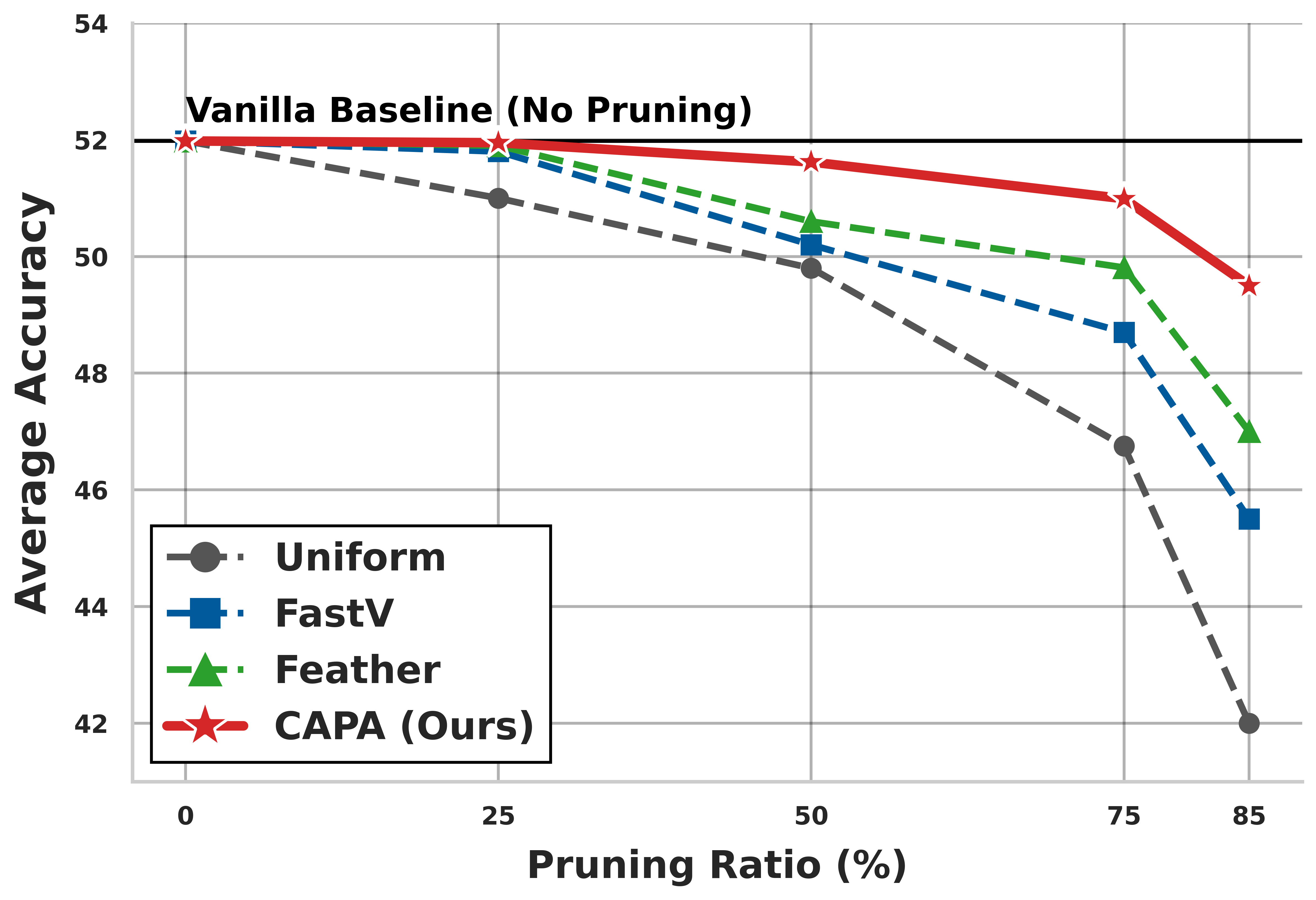}
\vspace{-4mm}
\caption{Average accuracy across six benchmarks under varying pruning ratios.
While baselines exhibit rapid degradation at high sparsity levels, \textbf{CAPA} maintains near-lossless performance up to 85\% sparsity.}
\vspace{-3mm}
\label{fig:robustness}
\end{figure}

\paragraph{FFN Approximation Efficiency.}
Tab.~\ref{tab:ffn_results} delineates the impact of our approximation strategies, both in isolation and when coupled with token pruning. We first observe that naively removing FFNs (\textit{Vanilla + Skip FFN}) leads to significant performance degradation, confirming that these layers retain non-negligible functional importance. However, replacing them with our proposed linear approximation (\textit{Vanilla + Hadamard product}) successfully recovers the majority of this performance drop, validating our hypothesis that the transformation in these layers is predominantly linear. When integrated with contribution-aware pruning, simply skipping FFN layers (\textit{CAPA with Skip FFN}) proves insufficient; in contrast, the fully combined framework (\textit{CAPA Combined}) which pairs token pruning with the Hadamard product approximation achieves the optimal trade-off between computational efficiency and model accuracy.

\begin{figure}[t]
\centering
\includegraphics[width=0.85\linewidth]{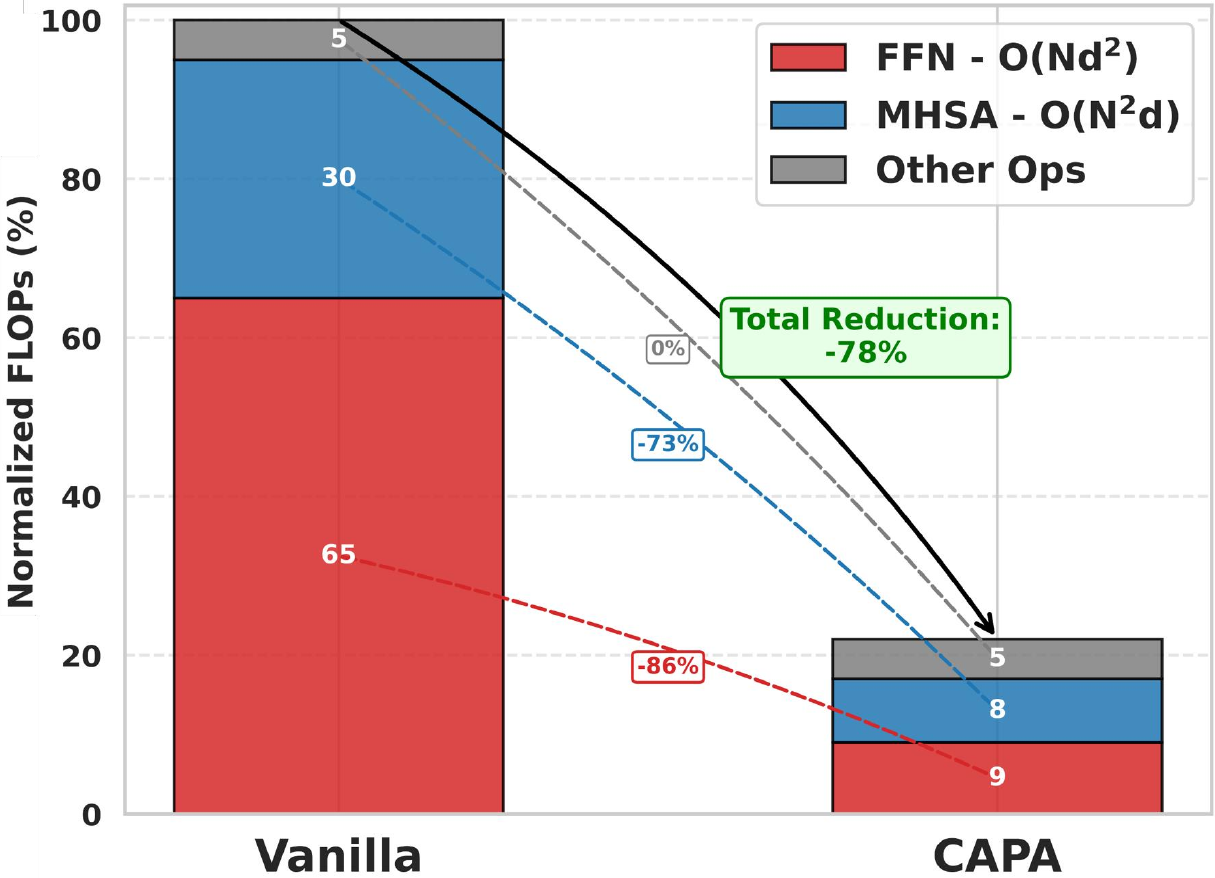}
\vspace{-4mm}
\caption{Theoretical inference FLOPs breakdown across model components. By identifying and approximating redundant FFNs (red), CAPA effectively removes the dominant $\mathcal{O}(Nd^2)$ cost with 78\% reduction.}
\vspace{-5mm}
\label{fig:flops}
\end{figure}

\subsection{Robustness and Efficiency Analysis}
We further analyze the trade-off between computational efficiency and model performance. Fig. \ref{fig:robustness} illustrates the robustness of CAPA across varying pruning ratios. Unlike prior methods such as FastV \cite{chen2024image} and Feather \cite{endo2025feather}, which experience precipitous performance drops beyond 50\% pruning, CAPA maintains near-vanilla accuracy ($\approx 52.0$ average score) even at an aggressive 75\% pruning ratio. This stability validates that Attention Contribution is a superior proxy for token saliency than raw attention scores. 

In addition, we quantify the computational gains in Fig. \ref{fig:flops}. Standard LVLM inference is dominated by FFNs, which account for approximately 65\% of total FLOPs. By replacing these dense matrix multiplications with our lightweight approximation in redundant layers, CAPA virtually eliminates this bottleneck (red region). When combined with the quadratic speedup from token pruning (blue region), our framework achieves a total FLOPs reduction of 78\%, offering a decisive efficiency over baselines that rely on token pruning alone.

\section{Conclusion}
In this paper, we introduced CAPA, an efficient framework for LVLM inference. By shifting pruning from attention scores to attention contribution magnitude and approximating redundant FFNs via lightweight Hadamard products, CAPA preserves critical visual information while significantly reducing computation. Our results highlight the need for visual-specific optimization distinct from text to maximize efficiency without sacrificing capability.

\section{Limitations}
While our post-hoc strategies contribution-aware pruning and FFN approximation yield substantial efficiency gains, they are not a complete solution. Post-hoc methods operate on fixed, pretrained backbones and therefore cannot fully exploit the benefits of jointly learned token selection and model computation. A promising direction for future work is the design of learned controllers that adaptively decide, for each input and at each layer, how many visual tokens to retain (or how much computation to allocate). Such layer-wise and data-dependent policies could enable finer-grained trade-offs between computation and accuracy than static pruning heuristics. 

Another important avenue is architectural: developing vision–language backbones in which the standard \texttt{FFN} is replaced or reparameterized by more efficient modules that are specifically tailored to visual tokens (\textit{e.g.,} lightweight element-wise transforms, low-rank or conditional linear layers, or mixture-of-expert style blocks). Jointly optimizing token-retention policies and such efficient FFN alternatives during training may yield models that are both faster and more robust than what post-hoc modifications can achieve.

Finally, future studies should evaluate these ideas across diverse benchmarks and distribution shifts, investigate calibration and interpretability of learned policies, and quantify the trade-offs between dynamic sparsity, latency, and downstream task performance.


\bibliography{custom}

\appendix

\clearpage
\newpage

\section{Appendix}
\subsection{Computational Complexity Analysis}
\label{sec:appendix_complexity}

In this section, we provide a theoretical analysis of the computational efficiency gains introduced by our proposed framework. We examine the Floating Point Operations (FLOPs) reduction achieved through two primary mechanisms: (1) Hadamard Product Approximation for redundant FFN layers, and (2) Contribution-Aware Vision Token Pruning.
\subsubsection{Efficiency of Hadamard FFN Approximation}
Standard LVLMs such as LLaVA and QwenVL utilize SwiGLU-based Feed-Forward Networks (FFNs). Let $d$ denote the hidden dimension size and $d_{ff}$ denote the intermediate dimension size (typically $d_{ff} \approx 4d$ or higher).
\paragraph{Baseline Complexity (Standard FFN).}
A standard SwiGLU FFN consists of three dense matrix multiplications (Gate, Up, and Down projections). For a single token, the computational cost is:
\begin{equation}
    \text{FLOPs}_{\text{FFN}} = 2 \cdot (3 \cdot d \cdot d_{ff}) = 6 d d_{ff}
\end{equation}
where the factor of 2 accounts for the multiply-accumulate operations.
\paragraph{Approximated Complexity (Hadamard Product).}
Our proposed approximation replaces these matrix operations with a single element-wise Hadamard product $\mathbf{x} \odot \boldsymbol{\alpha}$, where $\boldsymbol{\alpha} \in \mathbf{R}^d$. The cost for this operation is:
\begin{equation}
    \text{FLOPs}_{\text{Had}} = d
\end{equation}
\paragraph{Reduction Analysis.}
The reduction factor $\mathcal{R}$ for a single approximated layer is:
\begin{equation}
    \mathcal{R} = \frac{\text{FLOPs}_{\text{FFN}}}{\text{FLOPs}_{\text{Had}}} = \frac{6 d d_{ff}}{d} = 6 d_{ff}
\end{equation}
Given that $d_{ff}$ is in the order of thousands (e.g., $d_{ff}=11,008$ for LLaMA-2-7B), the computational cost of the approximated layer becomes negligible ($\approx 0.01\%$ of the original cost). For a model with $L$ total layers where $L_{approx}$ layers are selected for approximation, the total FFN FLOPs reduction is proportional to $\frac{L_{approx}}{L}$.
\subsubsection{Impact of Contribution-Aware Token Pruning}
We reduce the visual token sequence length by pruning 75\% of tokens based on their attention contribution scores. Let $N_{img}$ be the number of image tokens and $N_{txt}$ be the number of text tokens. The total sequence length is $N = N_{img} + N_{txt}$.
Let $\rho = 0.75$ represent the pruning rate applied to image tokens. The reduced sequence length is effective for all layers subsequent to the pruning stage. The effective sequence length becomes:
\begin{equation}
    N' = (1 - \rho)N_{img} + N_{txt}
\end{equation}
\paragraph{Reduction in Linear Projections (FFNs and QKV).}
Linear layers (FFNs and Attention projections) have a complexity of $\mathcal{O}(N \cdot d^2)$. The FLOPs reduction is linear with respect to the token count:
\begin{equation}
    \text{Speedup}_{\text{Linear}}
    = \frac{N}{N'}
    \approx \frac{1}{1 - \rho}
    \quad (N_{\mathrm{img}} \gg N_{\mathrm{txt}}).
\end{equation}
For $\rho=0.75$, this yields a theoretical \textbf{4$\times$ reduction} in FLOPs for all dense layers operating on the visual sequence.
\paragraph{Reduction in Attention Mechanism.}
The self-attention mechanism has a complexity of $\mathcal{O}(N^2 \cdot d)$. Since the complexity is quadratic with respect to sequence length, the pruning yields a significantly higher speedup:
\begin{equation}
    \text{Speedup}_{\text{Attn}} = \left( \frac{N}{N'} \right)^2 \approx \frac{1}{(1 - \rho)^2}
\end{equation}
With $\rho=0.75$, the attention mechanism theoretically becomes \textbf{16$\times$ faster} for the visual component.
\paragraph{Total Theoretical Reduction.}
Combining both strategies, our framework achieves a compounding efficiency gain. The Hadamard product approximation eliminates the $\mathcal{O}(d^2)$ cost of FFNs in redundant layers entirely, while token pruning reduces the $N$ coefficient for all remaining active layers. This dual approach ensures that we attack the computational bottleneck from both the \textit{width} (hidden dimension complexity) and the \textit{length} (sequence complexity) of the model.
\begin{figure*}[h]
    \centering
    \includegraphics[width=1.0\linewidth]{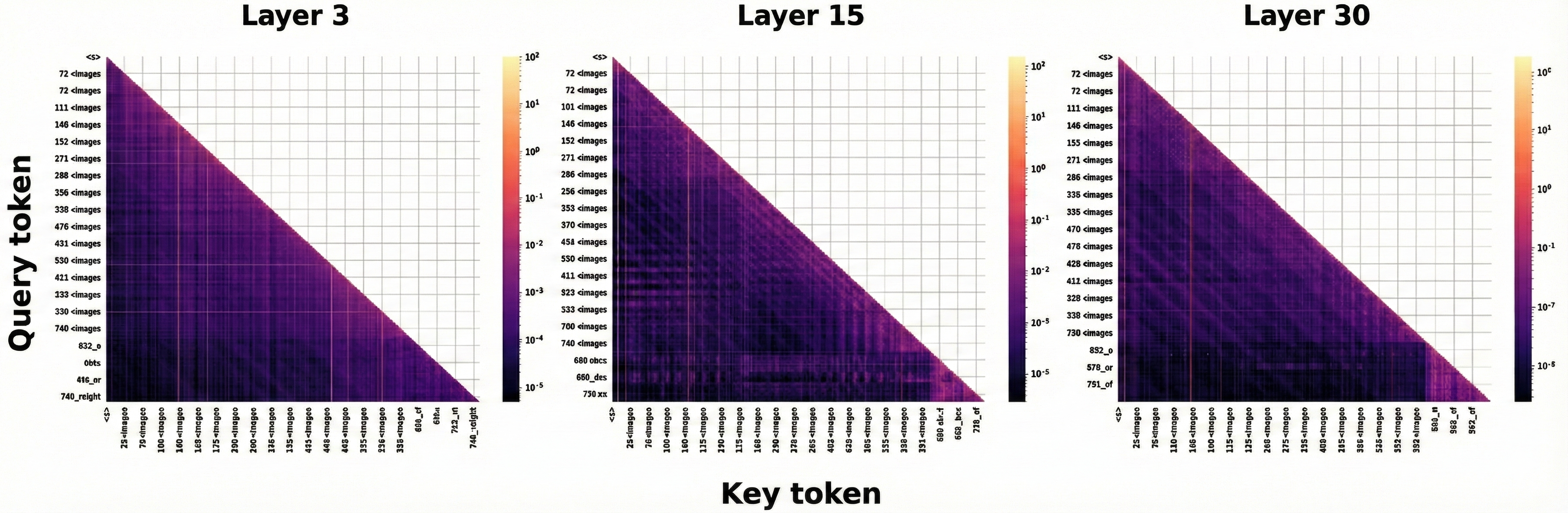}
    \caption{The layer-wise evolution of average attention contribution originating from query vectors and directed towards key vectors for all the tokens generated from a prompt "Describe the image in detail." across all layers of the in LLaVA-1.5.\textbf{Note: The contribution magnitude is plotted on a logarithmic scale}}
    \label{fig:contribution_evolution}
\end{figure*}
\subsection{Derivation of Optimal Hadamard Scaling}
In this section, we provide the complete mathematical derivation for the closed-form solution used to approximate the Feed-Forward Network (FFN) layers. As described in Section 4.2, our objective is to replace the computationally expensive dense matrix operations in redundant layers with a lightweight, element-wise Hadamard product. We achieve this by finding a learned scaling vector $\alpha \in \mathbb{R}^d$ that minimizes the reconstruction error between the approximated output and the true residual output of the FFN block.
\paragraph{Problem Formulation}
Let $x \in \mathbb{R}^d$ denote the input hidden state to a specific layer, and let $y \in \mathbb{R}^d$ denote the target output of the residual block, defined as $y = x + \text{FFN}(x)$. We seek an optimal scaling vector $\alpha$ such that the element-wise product $\hat{y} = x \odot \alpha$ approximates $y$ with minimal error.
We formulate this as an Ordinary Least Squares (OLS) regression problem. Since the Hadamard product operates independently on each feature dimension, we can decompose the optimization into $d$ independent scalar problems. For a specific feature dimension $k \in \{1, \dots, d\}$, we aim to minimize the sum of squared errors over a calibration dataset of $N$ samples.
The objective function $J(\alpha_k)$ for the $k$-th dimension is defined as:
\begin{equation}
    J(\alpha_k) = \sum_{n=1}^{N} (\alpha_k x_{n,k} - y_{n,k})^2
\end{equation}
where $x_{n,k}$ and $y_{n,k}$ represent the values of the $k$-th feature for the $n$-th sample in the calibration set.
\paragraph{Optimization via Gradient Analysis}
To find the global minimum for this convex objective function, we compute the partial derivative of $J(\alpha_k)$ with respect to the parameter $\alpha_k$ and set it to zero:
\begin{equation}
    \frac{\partial J}{\partial \alpha_k} = \sum_{n=1}^{N} 2(\alpha_k x_{n,k} - y_{n,k}) \cdot x_{n,k} = 0
\end{equation}
We can distribute the summation and factor out the constants to isolate $\alpha_k$:
\begin{equation}
    \sum_{n=1}^{N} (\alpha_k x_{n,k}^2 - x_{n,k} y_{n,k}) = 0
\end{equation}
\begin{equation}
    \alpha_k \sum_{n=1}^{N} x_{n,k}^2 - \sum_{n=1}^{N} x_{n,k} y_{n,k} = 0
\end{equation}
Rearranging the terms yields the closed-form solution for the optimal scalar $\alpha_k$:
\begin{equation}
    \alpha_k = \frac{\sum_{n=1}^{N} x_{n,k} y_{n,k}}{\sum_{n=1}^{N} x_{n,k}^2}
\end{equation}
This result confirms that the optimal scaling factor is simply the ratio of the cross-correlation between the input and target to the auto-correlation of the input.
\paragraph{Vectorized Implementation}
For efficient computation on GPUs, we vectorize this operation across all $d$ dimensions simultaneously. Let $X \in \mathbb{R}^{N \times d}$ and $Y \in \mathbb{R}^{N \times d}$ be the matrices containing the collected calibration statistics. The optimal vector $\alpha^*$ can be computed using element-wise operations:
\begin{equation}
    \alpha^* = \frac{\sum_{n=1}^{N} (x_n \odot y_n)}{\sum_{n=1}^{N} (x_n \odot x_n)}
\end{equation}
where $\odot$ denotes the element-wise product and the division is performed element-wise.
This analytical approach avoids the need for iterative gradient descent optimization. It allows us to calibrate the approximation parameters efficiently using only a small set of 500 samples, as the feature statistics converge rapidly to the optimal solution.
\subsection{Evolutionary Trajectory of Visual Information Flow}
It is crucial to analyze how the model's reliance on visual information evolves as data propagates through the layers. Fig.~\ref{fig:contribution_evolution} visualizes the average attention contribution originating from query vectors and directed towards key vectors for all the tokens generated from a prompt "Describe the image in detail." across all layers of the LLaVA-1.5 architecture.
This quantitative analysis reveals a distinct, progressive shift in the model's internal processing mechanism, characterized by three phases:
\paragraph{1. The Multimodal Fusion Phase (Early Layers).}
In the initial third of the network, we observe sustained, high-magnitude attention contributions from text to image tokens. During this stage, the model is actively engaged in grounding linguistic inputs within the visual context. The text tokens heavily query the visual features to resolve ambiguities and build an integrated multimodal representation.
\paragraph{2. The Abstraction Transition (Middle Layers).}
The middle layers mark a critical phase transition. The figure illustrates a precipitous drop in the average contribution score from text to image tokens. This indicates that the primary task of multimodal fusion is nearing completion. The essential semantic content from the image has been abstracted and integrated into the evolving textual representation. As a result, the raw visual tokens become less critical for immediate processing, though they may still be retained as fallback context.
\paragraph{3. The Deep Reasoning Phase (Late Layers).}
In the final layers, the attention contribution to image tokens nears zero. At this depth, the model operates almost exclusively on the fused, high-level textual representation. The generative process is driven by linguistic reasoning and next-token prediction dynamics, bearing strong resemblance to a pure large language model. The visual tokens at this stage which have any relevant information are Structural Anchors. This empirical observation provides the foundational justification for why aggressive pruning strategies even heuristic ones achieve near-lossless performance in the deepest layers of Large Multimodal Models.
An interesting thing to note here is that this phenomenon only occurs in vision tokens and not in text tokens highlighting the high entropy on vision tokens as compared to text tokens and the sharp decline in contribution in deeper layers empirically validates the hypothesis that the model's dependency on raw visual tokens diminishes significantly after initial multimodal fusion, rendering them redundant in the final stages of processing.
\subsection{Theoretical Analysis of Visual FFN Redundancy}
In this section, we provide a theoretical framework to explain the layer-wise FFN redundancy profile observed in our empirical analysis. Specifically, we address why visual tokens consistently exhibit high cosine similarity (linearity) in the intermediate layers of the network, contrasting sharply with the non-linear evolution of text tokens. We ground this behavior in the Information Saturation Hypothesis and the functional staging of Multimodal Large Language Models.
\subsubsection{The Read-Only Manifold Hypothesis}
We posit that following the initial multimodal fusion phase, visual tokens and text tokens occupy functionally distinct manifolds within the residual stream. The text generation process represents a dynamic system where the text state must evolve layer-by-layer to reduce entropy for next-token prediction. This necessitates significant non-linear FFN transformations to disentangle semantic features and perform reasoning.
In contrast, visual tokens in the intermediate layers serve primarily as conditioning constants or a static context for the text tokens. Once the visual features are projected into the semantic space of the language model during the initial layers, they must remain representationally stable. This stability is required to serve as a reliable addressable memory for the attention mechanism. If visual tokens were to undergo severe non-linear transformations via FFNs in every layer, the semantic alignment established in the early layers would drift, degrading the addressing system used by the text tokens to retrieve visual information.
To maintain this stability, visual tokens in the middle layers converge to fixed points of the layer function. For a visual token $x$, the update rule approximates an identity mapping where the FFN output approaches zero or acts linearly. This constraint explains the high cosine similarity observed in the middle layers, as the residual connection dominates the update.
\subsection{Functional Staging and Information Saturation}
The observed redundancy profile aligns with the three-stage processing hierarchy identified in recent probing studies \cite{yu2025multimodal}. We map the FFN redundancy dynamics to these functional stages:
\paragraph{Visual Grounding (Early Layers)} 
In the initial layers, the model aligns visual features with the embedding space of the language model. While some non-linearity is required here to project features into the correct subspace, the redundancy metric rises quickly. This suggests that the projection head and the first few transformer layers handle the bulk of this alignment, rapidly stabilizing the visual representation.
\paragraph{Context Absorption and Saturation (Middle Layers)}
During the intermediate layers, the text tokens actively query the visual tokens to resolve semantic references and integrate multimodal information. We hypothesize that this phase is characterized by Information Saturation. Once the text tokens have absorbed the necessary visual context via cross-modal attention, the visual tokens effectively become informationally saturated sources. They are no longer the target of processing but rather the static reference. Mathematically, as the gradient of information flow becomes unidirectional (from image to text), the utility of FFNs for updating visual tokens diminishes. The FFNs thus collapse to linearity, validating why approximating these layers results in minimal performance loss.
\paragraph{Decoupling and Decoding (Late Layers)}
We observe a decline in cosine similarity for visual tokens in the final layers. This corresponds to the decoding phase described by \cite{yu2025multimodal}, where the model shifts entirely to next-token prediction dynamics. At this stage, the attention mechanism often suppresses visual tokens as the model focuses on linguistic formatting and output generation. The divergence from linearity in these final layers likely reflects a semantic decoupling, where visual tokens are either transformed to disentangle task-specific features or simply drift due to a lack of attention constraints.
\end{document}